\documentclass{article}

\usepackage[final]{neurips_2023}

\usepackage[utf8]{inputenc} 
\usepackage[T1]{fontenc}    
\usepackage{hyperref}       
\usepackage{url}            
\usepackage{booktabs}       
\usepackage{amsfonts}       
\usepackage{nicefrac}       
\usepackage{microtype}      
\usepackage[dvipsnames]{xcolor}         
\usepackage{natbib}
\usepackage{amsmath}
\usepackage{amssymb}
\usepackage{cleveref}
\usepackage{graphicx}

\title{Hazards from Increasingly Accessible Fine-Tuning of Downloadable Foundation Models}

\author{%
  Alan Chan$^{*+}$\\
  Mila - Quebec AI Institute\\
  Université de Montréal\\
  \And
  Ben Bucknall$^{*\circ}$\\
  Independent\\
  \And
  Herbie Bradley\\
  UK AI Safety Institute \\
  University of Cambridge\\
  \And
  David Krueger\\
  University of Cambridge\\
}

\begin{document}

\maketitle

\def\thefootnote{*}\footnotetext{Equal contribution.}
\def\thefootnote{+}\footnotetext{Correspondence to \texttt{alan.chan@mila.quebec}.}
\def\thefootnote{$\circ$}\footnotetext{Work completed as an intern with Alan Chan and David Krueger.}
\renewcommand{\thefootnote}{\arabic{footnote}}

\begin{abstract}
Public release of the weights of pretrained foundation models, otherwise known as downloadable access \citep{solaiman_gradient_2023}, enables fine-tuning without the prohibitive expense of pretraining. Our work argues that increasingly accessible fine-tuning of downloadable models may increase hazards. First, we highlight research to improve the accessibility of fine-tuning. We split our discussion into research that A) reduces the computational cost of fine-tuning and B) improves the ability to share that cost across more actors. Second, we argue that increasingly accessible fine-tuning methods may increase hazard through facilitating malicious use and making oversight of models with potentially dangerous capabilities more difficult. Third, we discuss potential mitigatory measures, as well as benefits of more accessible fine-tuning. Given substantial remaining uncertainty about hazards, we conclude by emphasizing the urgent need for the development of mitigations.
\end{abstract}

\section{Introduction}
A strong consensus exists in software engineering and cryptography that openness is the best guarantee of security \citep{anderson_open_2007, swire_model_2004, diffie_export_2007}. Some have argued that openness best addresses the risks of foundation models \citep{kapoor_three_2023, kapoor_licensing_2023, howard_ai_2023, mostaque_importance_2023}, while others argue to the contrary that open release strategies can facilitate misuse \citep{anderljung_protecting_2023,hazell_large_2023,brundage_toward_2020}. 

We focus on the risks of increasingly accessible fine-tuning of downloadable models, where by \textbf{downloadable model}, we mean a model whose pretrained weights are publicly available for download \citep{solaiman_gradient_2023}.
We also refer to this practice as \textbf{downloadable release} or providing \textbf{downloadable access}. Through fine-tuning, downloadable release facilitates the addition of capabilities to a model without the prohibitive expense of pretraining. Moreover, fine-tuning of downloadable models bypasses potential restrictions and filters which may be present on APIs that facilitate the fine-tuning of non-downloadable models, although such filters may be easy to circumvent \citep{qi_fine-tuning_2023}. We focus on downloadable access because inference and fine-tuning code is usually available in tandem with the weights or otherwise straightforward to implement---though many of our arguments also apply to potential hazard from the fine-tuning of \textit{non-downloadable models} (see Appendix~\ref{app:non-downloadable}).

We argue that increasingly accessible fine-tuning of downloadable models will likely increase hazard. \textbf{Risk} is given by the product of hazard, exposure, and vulnerability \citep{hendrycks_x-risk_2022}, where \textbf{hazard} is the severity and prevalence of sources of harm, \textbf{exposure} is how exposed we are to hazards, and \textbf{vulnerability} is our susceptibility to their damaging effects. We concentrate on hazard in this work given length limitations, but plan to extend our analysis in future work.

Our specific contributions are as follows. First, we highlight ongoing research areas that improve the accessibility of model fine-tuning. We split our discussion into research developments that A) reduce the costs of fine-tuning a model and B) facilitate the sharing of such costs across more actors. Second, we analyze the impact of these changes to the cost of fine-tuning on hazard, focusing on non-state, malicious actors and difficulties of imposing oversight. We conclude with a discussion of potential mitigation strategies and benefits of accessible fine-tuning.

\section{Changes to the Accessibility of Fine-Tuning}\label{sec:ft-change}
Here, we discuss research areas that increase the accessibility of fine-tuning for downloadable models. We split our discussion into A) reductions in the cost to obtain a given capability level and B) improvements in the ability to distribute said cost across more people, referred to as cost reduction and improved cost sharing, respectively. In \Cref{sec:frontier-development}, we discuss the impact of A) and B) relative to frontier development.


\subsection{Cost Reduction}

We first consider research to reduce the computational costs of attaining a given capability level.

\subsubsection{Improved Algorithms}
Cost reductions may be gained from the use of more efficient algorithms that directly reduce the number of floating-point operations (FLOP) or amount of on-chip memory required to attain a given performance level, while keeping the training data and model architecture constant.

Classical SGD, since it avoids storing the optimizer state variables used by Adam and other higher-order methods, can significantly reduce the required memory for fine-tuning. However, it is more brittle and requires significant time and compute spent tuning the learning rate.
Zeroth-order optimizers, which avoid 
the large memory requirements of backpropogation, may be promising \citep{malladi_fine-tuning_2023}.

Parameter offloading \citep{pudipeddi_training_2020, ren_zero-offload_2021} methods do not change total memory requirements but rather the GPU memory required, which is more costly than RAM or CPU memory. Such methods work by storing the model outside of the GPU memory (such as in RAM or CPU memory), and retrieving only individual layers' parameters from this external memory during training.

Finally, we also include novel training objectives.
For example, \citet{tay_transcending_2022} introduce a novel mixture of training objectives that purport to provide a 2x compute saving over standard pretraining objectives.

\subsubsection{Better Synthetic Data}
Secondly, we consider the use of synthetic data as a method of reducing the cost of constructing specialized fine-tuning datasets. There is currently uncertainty surrounding the utility and potential adverse side-effects of training on synthetic data. Some authors claim that it harms model performance \citep{alemohammad_self-consuming_2023, shumailov_curse_2023, martinez_combining_2023, martinez_towards_2023}, while others claim that it can boost performance \citep{huang_large_2022,li_textbooks_2023,azizi_synthetic_2023}.

Instruction-tuning \citep{taori_alpaca_2023} may also be considered as a synthetic data method. Many have argued that instruction tuning is a cheap way of drastically improving the output quality of smaller base models by fine-tuning them on example outputs from a larger and more capable model \citep{taori_alpaca_2023, mukherjee_orca_2023, chiang_vicuna_2023, dubois_alpacafarm_2023}. However, others have argued that such methods only improve the appearance of model outputs whilst having little effect on the models' underlying quality and factuality \citep{gudibande_false_2023, wang_how_2023}.

While there may not currently be consensus on the effectiveness of synthetic fine-tuning data, 
it is plausible that the effectiveness of synthetic data will increase over time. If so, the costs of producing high-quality, specialized fine-tuning datasets for specific model capabilities could be greatly reduced.

\subsubsection{Parameter-Efficient Fine-Tuning}
Parameter-efficient fine-tuning methods aim to reduce the number of parameters that are fine-tuned while retaining model performance, thereby reducing the total FLOP required. These methods include prompt- and prefix-tuning \citep{liu_p-tuning_2022, liu_gpt_2021, lester_power_2021, qin_learning_2021, li_prefix-tuning_2021}, adapters \citep{rebuffi_learning_2017, houlsby_parameter-efficient_2019, liu_few-shot_2022}, and sparse fine-tuning \citep{guo_parameter-efficient_2021, ben-zaken_bitfit_2022, xu_raise_2021}. For example, \citet{hu_lora_2021} proposed Low-Rank Adaptation (LoRA), a method that applied low-rank matrix decomposition to reduce the number of fine-tuning parameters by four orders of magnitude when applied to GPT-3 175B, while appearing to perform on-par with fine-tuning all model parameters.

\subsubsection{Quantization}
Quantization reduces the floating-point precision of a model's weights while retaining performance, thus reducing the compute and memory footprint during inference and/or training. \citet{dettmers_llmint8_2022} introduce a mixed-precision quantization method that allows models with 2-3x more parameters to be run on given hardware, when compared to a standard 16-bit precision baseline, for a negligible increase in inference time. \citet{dettmers_qlora_2023} further extend this work, combining quantization with LoRA \citep{hu_lora_2021} to reduce the memory requirements of fine-tuning 15-fold.

\subsection{Improved Cost Sharing}
Cost-sharing methods allow one to distribute the cost of fine-tuning over a number of actors. Such methods enable collectives of actors to collaborate on fine-tuning, where each individual only has to pay a small percentage of the total cost involved. 
While both approaches considered below may \emph{increase} the total cost of fine-tuning, sharing this increased cost amongst enough individuals would mean a net cost reduction for each participating individual.

\subsubsection{Decentralized Training}
Decentralized training methods aim to allow efficient training on multiple devices, where these devices are assumed to be connected only through slow or low-bandwidth communication channels \citep{together_neurips_2022, together_neurips_2022-1}. Methods that are feasible under such an assumption would permit the training of AI systems on geographically distributed networks owned by multiple individuals, thus alleviating the need for large, centrally-owned datacenters. A related area of research is federated learning, although it is usually motivated by data privacy concerns \citep{khan_federated_2021, liu_federated_2021}.

Two predominant methods currently are efficient task scheduling and allocation \citep{yuan_decentralized_2023, ryabinin_swarm_2023, borzunov_petals_2023}, and compression of gradients \citep{dettmers_8-bit_2016} or activations \citep{wang_fine-tuning_2023} to reduce the information transferred between devices. As an example of the former, \citet{yuan_decentralized_2023} propose an algorithm for scheduling and allocation of training tasks among distributed devices that leaves the convergence dynamics unchanged. While their method results in a $1.7$-$3.5\times$ greater absolute time cost when compared to that of training in state-of-the-art centralized datacenters, the training took place on a distributed network that was $100\times$ slower.

\subsubsection{Composing Trained Models}
Cost sharing may also be achieved through the post-hoc composition of independent fine-tuned models, developed on top of a shared base model, resulting in a composite model that is more capable than any of its constituents \citep{li_branch-train-merge_2022, gururangan_scaling_2023, jang_exploring_2023}. For example, \citet{ponti_parameter_2020} and \citet{ansell_composable_2022} combine two fine-tuned models, trained on language $L$ and task $T$ respectively, to create a model that can perform task $T$ in language $L$. It might be cheaper for each individual to fine-tune for one component ability rather than for one individual to fine-tune for the composed ability, especially if fine-tuning data for the composed ability are not available.

\section{Impacts on Hazard}\label{sec:risks}
We argue that both A) cost reduction and B) improved cost sharing may increase hazard. In \Cref{sec:alg-progress}, we address the implications of algorithmic progress on our arguments. In \Cref{sec:mitigations}, we discuss potential mitigations.

Let $X$ denote some capability. 
In the following, we use the phrase \textit{fine-tuning a model to achieve $X$} to refer to the act of fine-tuning a model to obtain the capability $X$ or some threshold thereof. Let $A$ denote the actor that wishes to fine-tune a model to achieve $X$.

\subsection{Malicious Use Cost Reduction}\label{sec:risk-cost-reduction}
We claim that cost reduction may increase hazard through facilitating malicious use. As an illustrative example, we will consider $X = \text{execute chemical weapons attacks}$ and $A = (\text{small, non-state groups who wish to do } X)$. Our arguments will also apply to small, nonstate groups who want to carry out other kinds of attacks, as long as data are available to fine-tune a model to help with lowering the barrier to such attacks, or with coaching actors attempting to carry them out. Note that our argument is more applicable to fine-tuning of next-generation models, given that current-generation models do not seem to be helpful for our example of $X$ \citep{gopal_will_2023}.

\begin{figure}[h]
    \centering
    \includegraphics[width=0.65\textwidth]{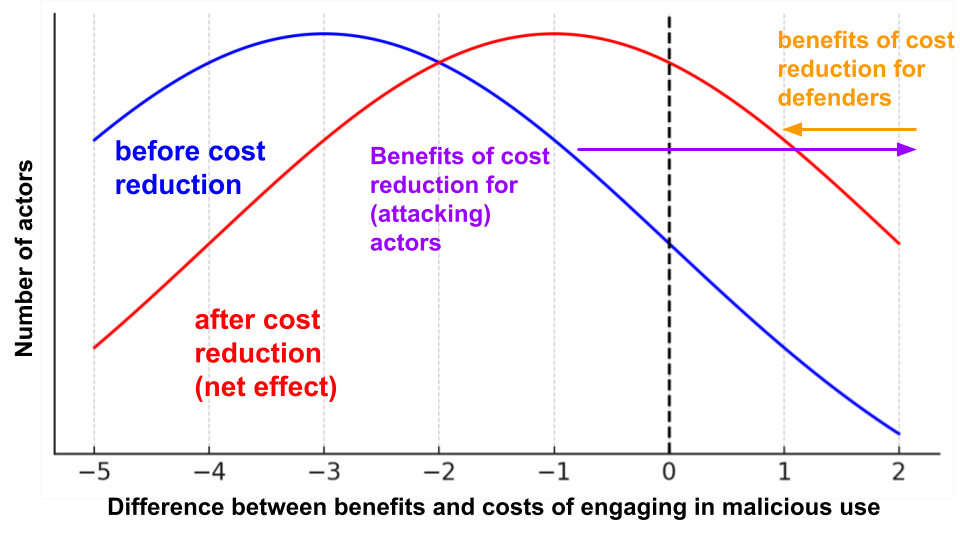}
    \caption{We illustrate our argument for the impact of cost reduction (reducing the computational cost of fine-tuning to a particular capability level). Each curve represents a distribution of actors according to their net benefit of engaging in malicious use, which in our illustrative example is carrying out a chemical attack. The \textbf{\textcolor{red}{red}} curve is shifted to the right of the \textbf{\textcolor{blue}{blue}} curve because of the net effect of cost reduction. Determining the net effect involves assessing the benefits of cost reduction to \textbf{\textcolor{orange}{defenders}} who wish to deter the actors, as well as the benefits of cost reduction for the \textbf{\textcolor[HTML]{9901ff}{attackers}}. We explain this argument in \Cref{sec:risk-cost-reduction}. } 
    \label{fig:cost-reduction}
\end{figure}

We need to show that more accessible fine-tuning shifts the cost-benefit considerations for $A$ in favour of malicious use. A subtlety here is that $A$ represents a class of actors, all with potentially different motivations and background costs. We are not arguing that every actor represented in $A$ will engage in malicious use; we argue that on net we should expect more actors to engage in malicious use. \Cref{fig:cost-reduction} provides an illustrative diagram for our claimed effect of cost reduction.

At a high level, the argument is as follows; we provide a more complete argument in \Cref{app:cost-reduction} due to space limitations. Actors in $A$ are assumed to have the motivation to engage in malicious use (in our example, chemical weapon attacks). We suppose that the benefits of carrying out such attacks do not depend upon the accessibility of fine-tuning; rather, they depend upon background motivations. We are left to analyze the effect of accessible fine-tuning on the costs of carrying out an attack. On the one hand, more accessible fine-tuning reduces $A$'s cost because `coaching' from fine-tuned models may lower the barrier for $A$ to engage in malicious use. On the other hand, more accessible fine-tuning could increase $A$'s cost because it may enable other actors, such as states, to retaliate against $A$, monitor $A$, or otherwise prevent $A$ from carrying out such attacks. We argue in \Cref{app:cost-reduction} that the net effect is to reduce $A$'s cost of engaging in malicious use, as we illustrate in \Cref{fig:cost-reduction}.

A final point concerns the severity of harm. Cost reduction could enable other actors to improve defenses such that even if more attacks occur, the total impact may be reduced. For example, cost reduction could enable more effective research on automated vulnerability finding in cybersecurity, helping defend against future AI assisted attacks. There remains substantial uncertainty about how the offence-defence balance scales with capabilities \citep{garfinkel_how_2019}, especially in specific domains such as cybersecurity. 

\subsection{Improved Cost Sharing}
We will argue that improved cost sharing may lead to an increase in the number of models with potentially dangerous capabilities or misalignment issues. While this increase itself may not cause problems given sufficient oversight, we will argue that in the absence of additional measures, current oversight practices will be insufficient. Harm may result from either misuse or misalignment. 

First, decentralized or distributed training methods may facilitate the construction of more capable models. Greatly improved decentralized training procedures could pull together a massive amount of disparate compute resources, exceeding those owned by even the largest AI labs. If decentralized training occurs at a comparable or greater scale to the most expensive centralized training runs, scaling laws suggest that the resulting models may have abilities that are at least comparable to the frontier, all other things equal. Moreover, composition of models trained to expert level on specific domains, such as code or biology, could lead to a single model that is generally capable across the constituent domains---although we note that model composition on this scale is still highly speculative.

The key point is not that cost sharing will result in models that significantly push the frontier, but that some of the resulting models, by virtue of criteria such as generality or the amount of compute used to train them (from decentralized training), will need to be evaluated for dangerous capabilities or misalignment problems  \citep{shevlane_model_2023}. There are likely to be actors who wish to build such capable models for legitimate reasons, such as the wish to customize a model for their own purposes, or simply the desire to create a downloadable model and make it available for others.

If we are able to robustly evaluate the resulting models for dangerous capabilities and misalignment as part of a potential pre-deployment evaluation regime, there would be no increase in hazard. However, model risk evaluation is an ongoing and imperfect science---robust evaluation methods are still lacking. Even if there were a mandatory evaluation scheme  \citep{anderljung_frontier_2023} applied to all models, such a scheme does not guarantee coverage of all significant risks. 
Moreover, evaluation methods for dangerous capabilities and misalignment tend to be expensive in terms of both time and money. For instance, expert red-teamers across a wide variety of domains may need to be hired to test the model for an extended period of time~\citep{openai_gpt-4_2023}. It seems difficult to impose these costs on decentralized developers, who may even be anonymous. A dedicated organization may perform evaluations of such models, but it may be difficult to keep up with a potential profusion of models to test---although this depends on the level of an evaluation threshold in terms of capabilities or compute. Advances in automatic evaluation could help, but such work is still preliminary \citep{perez_discovering_2022}.

\subsection{Potential Mitigation 
Strategies}\label{sec:mitigations}
A number of active research directions aim to mitigate the hazards we have identified. Although such work is still early, it would ideally minimize the hazards from fine-tuning while maintaining the benefits that we discuss later in \Cref{sec:benefits}.

First, one could try to make it more difficult to fine-tune a pretrained model for a particular task. For example, \citet{henderson_self-destructing_2023} provide a proof-of-concept to make fine-tuning a pretrained model for a designated harmful task as difficult as training a randomly initialized model from scratch. Such methods may make it more difficult to build models with dangerous capabilities and therefore help to address the hazards from cost reduction and cost sharing.

Second, machine unlearning \citep{bourtoule_machine_2020} could remove memorized information in potentially harmful domains, which could be exploited by malicious actors. For instance, removing information about chemical synthesis  from training datasets may be desirable.

Further research and development of safety approaches that are robust to downstream model modification, including fine-tuning, is much needed given the fragility and ease-of-removal of current, largely RLHF-based, approaches \citep{qi_fine-tuning_2023}.

\section{Benefits of Accessible Fine-Tuning}\label{sec:benefits}
We now assess some potential benefits to an increase in the accessibility of fine-tuning.
Firstly, the ability to fine-tune downloadable models is crucial for academic and independent research aimed at better understanding contemporary AI systems and improving their safety. A reduction in the cost of performing fine-tuning would enable lesser-resourced academic labs and independent researchers to conduct such research. For example, \citet{berglund_reversal_2023} relied on the ability to fine-tune for their investigation of the `reversal curse', that is, the inability of language models to infer statements of the form \textit{`A is B'} despite \textit{`B is A'} appearing in the model's fine-tuning data. 

Secondly, accessible fine-tuning facilitates the downstream adaption of AI models to novel use-cases, for which upstream model developers would not have the capacity to provide. For example, accessible fine-tuning has enabled the adaptation of language models to underrepresented or minority languages. While it may be unrealistic for a centralized AI developer to provide a different version of a language model for potentially hundreds of languages, cost reduction and improved cost sharing may allow open-source collectives to fine-tune a downloadable model to function in their local language. 


Broader access to capable models, facilitated in part by accessible fine-tuning, may allow society to avoid unsustainable power differentials between those that do and do not have direct access to the most capable AI models \citep{howard_ai_2023}. By default, wealthy entities like large corporations and states will be able to adapt foundation models for their own agendae, as well as exert disproportionate influence over the continued development of AI and its integration in society.
Such entities might use these models to influence information flows, shaping what the public believes for their own benefit. Broad public access to building similarly powerful models could resist this tendency. This access could enable civil society to fine-tune its own models to, for example, fact-check media or generally serve as public advocates.

\section{Related Work}
Many recent works have investigated downloadable release \citep{sastry_beyond_2021,liang_time_2022,solaiman_release_2019,ovadya_reducing_2019,whittlestone_tension_2020}. \citet{seger_open-sourcing_2023} evaluate the misuse and proliferation risks of downloadable release and propose methods to capture its benefits while avoiding the pitfalls. Some works have highlighted that openness is a spectrum \citep{shevlane_structured_2022,solaiman_gradient_2023}. 
Other works have analyzed conditions under which downloadable release is positive. \citet{kapoor_licensing_2023,kapoor_three_2023, phang_eleutherai_2022} argue that the safety of AI systems is best served through open study of their failures. \citet{shevlane_offense-defense_2020} study the impact of public AI research on misuse. \citet{anderljung_protecting_2023} argue that certain restrictions on AI capabilities may be warranted to reduce misuse. Other research connects open release to power. \citet{seger_democratising_2023} outline different ways to think about the democratization of AI; downloadable release most closely aids democratization of development and use. \citet{widder_open_2023} argue that current practices of openness fail to enable oversight of and democratic access to AI systems. 

While we focus on access to capabilities, \citet{davidson_ai_nodate} argue that post-training enhancements can push the frontier. They show that enhancements such as scaffolding and tool integration can significantly enhance capabilities, at a low computational cost relative to pretraining. 

\section{Conclusion}

We have argued that increasingly accessible fine-tuning of downloadable models may increase hazard. We highlighted several ongoing areas of research that either reduce the cost of fine-tuning or allow the cost to be better shared among a group of actors. We then argued that A) cost reduction may make it easier for small non-state actors to cause harm, using the production of chemical weaponry as an illustrative example; and B) cost sharing may make it more difficult to impose oversight on models with potentially dangerous capabilities. We highlighted some initial directions to mitigate these hazards, in addition to potential benefits of more accessible fine-tuning.

There are many remaining uncertainties to be addressed in future work.
Firstly, other potential concerns of downloadable models, such as difficulties with attributing liability, should be investigated. 
Secondly, the availability of fine-tuning data remains a significant uncertainty in our analysis. Thirdly, offense-defense balances remain uncertain. Even if there are more instances of harm, those harms may have a smaller total impact because of improvements in defense. 

An ideal scenario would involve minimizing the hazards while maintaining most of the potential benefits. Since the hazards have not yet materialized, we should allocate more research effort to i) studying when the hazards might materialize and ii) developing mitigations, such as those that make fine-tuning difficult on certain tasks or for certain capabilities. 

\begin{ack}
We are grateful to the following people for helpful conversations and feedback relating to this work: 
Max Kaufmann, Anson Ho, Usman Anwar, Shalaleh Rismani, Gabe Mukobi, Stella Biderman, Lennart Heim, and Elizabeth Seger.

BB received funding from the Berkeley Existential Risk Initiative for this work. 
\end{ack}

\bibliographystyle{plainnat}
\bibliography{refs}

\begin{thebibliography}{91}
\providecommand{\natexlab}[1]{#1}
\providecommand{\url}[1]{\texttt{#1}}
\expandafter\ifx\csname urlstyle\endcsname\relax
  \providecommand{\doi}[1]{doi: #1}\else
  \providecommand{\doi}{doi: \begingroup \urlstyle{rm}\Url}\fi

\bibitem[Alemohammad et~al.(2023)Alemohammad, Casco-Rodriguez, Luzi, Humayun,
  Babaei, LeJeune, Siahkoohi, and Baraniuk]{alemohammad_self-consuming_2023}
Sina Alemohammad, Josue Casco-Rodriguez, Lorenzo Luzi, Ahmed~Imtiaz Humayun,
  Hossein Babaei, Daniel LeJeune, Ali Siahkoohi, and Richard~G. Baraniuk.
\newblock Self-{Consuming} {Generative} {Models} {Go} {MAD}, July 2023.
\newblock URL \url{http://arxiv.org/abs/2307.01850}.
\newblock arXiv:2307.01850 [cs].

\bibitem[Anderljung and Hazell(2023)]{anderljung_protecting_2023}
Markus Anderljung and Julian Hazell.
\newblock Protecting {Society} from {AI} {Misuse}: {When} are {Restrictions} on
  {Capabilities} {Warranted}?, March 2023.
\newblock URL \url{http://arxiv.org/abs/2303.09377}.
\newblock arXiv:2303.09377 [cs].

\bibitem[Anderljung et~al.(2023)Anderljung, Barnhart, Korinek, Leung, O'Keefe,
  Whittlestone, Avin, Brundage, Bullock, Cass-Beggs, Chang, Collins, Fist,
  Hadfield, Hayes, Ho, Hooker, Horvitz, Kolt, Schuett, Shavit, Siddarth,
  Trager, and Wolf]{anderljung_frontier_2023}
Markus Anderljung, Joslyn Barnhart, Anton Korinek, Jade Leung, Cullen O'Keefe,
  Jess Whittlestone, Shahar Avin, Miles Brundage, Justin Bullock, Duncan
  Cass-Beggs, Ben Chang, Tantum Collins, Tim Fist, Gillian Hadfield, Alan
  Hayes, Lewis Ho, Sara Hooker, Eric Horvitz, Noam Kolt, Jonas Schuett, Yonadav
  Shavit, Divya Siddarth, Robert Trager, and Kevin Wolf.
\newblock Frontier {AI} {Regulation}: {Managing} {Emerging} {Risks} to {Public}
  {Safety}, September 2023.
\newblock URL \url{http://arxiv.org/abs/2307.03718}.
\newblock arXiv:2307.03718 [cs].

\bibitem[Anderson(2007)]{anderson_open_2007}
Ross Anderson.
\newblock Open and {Closed} {Systems} are {Equivalent} (that is, in an ideal
  world).
\newblock In \emph{Perspectives on free and open source software}, pages
  127--142. MIT Press, January 2007.
\newblock URL
  \url{https://www.research.ed.ac.uk/en/publications/open-and-closed-systems-are-equivalent-that-is-in-an-ideal-world}.

\bibitem[Ansell et~al.(2022)Ansell, Ponti, Korhonen, and
  Vulić]{ansell_composable_2022}
Alan Ansell, Edoardo Ponti, Anna Korhonen, and Ivan Vulić.
\newblock Composable {Sparse} {Fine}-{Tuning} for {Cross}-{Lingual} {Transfer}.
\newblock In \emph{Proceedings of the 60th {Annual} {Meeting} of the
  {Association} for {Computational} {Linguistics} ({Volume} 1: {Long}
  {Papers})}, pages 1778--1796, Dublin, Ireland, May 2022. Association for
  Computational Linguistics.
\newblock \doi{10.18653/v1/2022.acl-long.125}.
\newblock URL \url{https://aclanthology.org/2022.acl-long.125}.

\bibitem[Azizi et~al.(2023)Azizi, Kornblith, Saharia, Norouzi, and
  Fleet]{azizi_synthetic_2023}
Shekoofeh Azizi, Simon Kornblith, Chitwan Saharia, Mohammad Norouzi, and
  David~J. Fleet.
\newblock Synthetic {Data} from {Diffusion} {Models} {Improves} {ImageNet}
  {Classification}, April 2023.
\newblock URL \url{http://arxiv.org/abs/2304.08466}.
\newblock arXiv:2304.08466 [cs].

\bibitem[Bai et~al.(2022)Bai, Kadavath, Kundu, Askell, Kernion, Jones, Chen,
  Goldie, Mirhoseini, McKinnon, Chen, Olsson, Olah, Hernandez, Drain, Ganguli,
  Li, Tran-Johnson, Perez, Kerr, Mueller, Ladish, Landau, Ndousse, Lukosuite,
  Lovitt, Sellitto, Elhage, Schiefer, Mercado, DasSarma, Lasenby, Larson,
  Ringer, Johnston, Kravec, Showk, Fort, Lanham, Telleen-Lawton, Conerly,
  Henighan, Hume, Bowman, Hatfield-Dodds, Mann, Amodei, Joseph, McCandlish,
  Brown, and Kaplan]{bai_constitutional_2022}
Yuntao Bai, Saurav Kadavath, Sandipan Kundu, Amanda Askell, Jackson Kernion,
  Andy Jones, Anna Chen, Anna Goldie, Azalia Mirhoseini, Cameron McKinnon,
  Carol Chen, Catherine Olsson, Christopher Olah, Danny Hernandez, Dawn Drain,
  Deep Ganguli, Dustin Li, Eli Tran-Johnson, Ethan Perez, Jamie Kerr, Jared
  Mueller, Jeffrey Ladish, Joshua Landau, Kamal Ndousse, Kamile Lukosuite,
  Liane Lovitt, Michael Sellitto, Nelson Elhage, Nicholas Schiefer, Noemi
  Mercado, Nova DasSarma, Robert Lasenby, Robin Larson, Sam Ringer, Scott
  Johnston, Shauna Kravec, Sheer~El Showk, Stanislav Fort, Tamera Lanham,
  Timothy Telleen-Lawton, Tom Conerly, Tom Henighan, Tristan Hume, Samuel~R.
  Bowman, Zac Hatfield-Dodds, Ben Mann, Dario Amodei, Nicholas Joseph, Sam
  McCandlish, Tom Brown, and Jared Kaplan.
\newblock Constitutional {AI}: {Harmlessness} from {AI} {Feedback}, December
  2022.
\newblock URL \url{http://arxiv.org/abs/2212.08073}.
\newblock arXiv:2212.08073 [cs].

\bibitem[Ben-Zaken et~al.(2022)Ben-Zaken, Ravfogel, and
  Goldberg]{ben-zaken_bitfit_2022}
Elad Ben-Zaken, Shauli Ravfogel, and Yoav Goldberg.
\newblock {BitFit}: {Simple} {Parameter}-efficient {Fine}-tuning for
  {Transformer}-based {Masked} {Language}-models, September 2022.
\newblock URL \url{http://arxiv.org/abs/2106.10199}.
\newblock arXiv:2106.10199 [cs].

\bibitem[Berglund et~al.(2023)Berglund, Tong, Kaufmann, Balesni, Stickland,
  Korbak, and Evans]{berglund_reversal_2023}
Lukas Berglund, Meg Tong, Max Kaufmann, Mikita Balesni, Asa~Cooper Stickland,
  Tomasz Korbak, and Owain Evans.
\newblock The {Reversal} {Curse}: {LLMs} trained on "{A} is {B}" fail to learn
  "{B} is {A}", September 2023.
\newblock URL \url{http://arxiv.org/abs/2309.12288}.
\newblock arXiv:2309.12288 [cs].

\bibitem[Boiko et~al.(2023)Boiko, MacKnight, and Gomes]{boiko_emergent_2023}
Daniil~A. Boiko, Robert MacKnight, and Gabe Gomes.
\newblock Emergent autonomous scientific research capabilities of large
  language models, April 2023.
\newblock URL \url{http://arxiv.org/abs/2304.05332}.
\newblock arXiv:2304.05332 [physics].

\bibitem[Borzunov et~al.(2023)Borzunov, Baranchuk, Dettmers, Ryabinin, Belkada,
  Chumachenko, Samygin, and Raffel]{borzunov_petals_2023}
Alexander Borzunov, Dmitry Baranchuk, Tim Dettmers, Max Ryabinin, Younes
  Belkada, Artem Chumachenko, Pavel Samygin, and Colin Raffel.
\newblock Petals: {Collaborative} {Inference} and {Fine}-tuning of {Large}
  {Models}, March 2023.
\newblock URL \url{http://arxiv.org/abs/2209.01188}.
\newblock arXiv:2209.01188 [cs].

\bibitem[Bourtoule et~al.(2020)Bourtoule, Chandrasekaran, Choquette-Choo, Jia,
  Travers, Zhang, Lie, and Papernot]{bourtoule_machine_2020}
Lucas Bourtoule, Varun Chandrasekaran, Christopher~A. Choquette-Choo, Hengrui
  Jia, Adelin Travers, Baiwu Zhang, David Lie, and Nicolas Papernot.
\newblock Machine {Unlearning}, December 2020.
\newblock URL \url{http://arxiv.org/abs/1912.03817}.
\newblock arXiv:1912.03817 [cs].

\bibitem[Bran et~al.(2023)Bran, Cox, Schilter, Baldassari, White, and
  Schwaller]{bran_chemcrow_2023}
Andres~M. Bran, Sam Cox, Oliver Schilter, Carlo Baldassari, Andrew~D. White,
  and Philippe Schwaller.
\newblock {ChemCrow}: {Augmenting} large-language models with chemistry tools,
  June 2023.
\newblock URL \url{http://arxiv.org/abs/2304.05376}.
\newblock arXiv:2304.05376 [physics, stat].

\bibitem[Brundage et~al.(2020)Brundage, Avin, Wang, Belfield, Krueger,
  Hadfield, Khlaaf, Yang, Toner, Fong, Maharaj, Koh, Hooker, Leung, Trask,
  Bluemke, Lebensold, O'Keefe, Koren, Ryffel, Rubinovitz, Besiroglu, Carugati,
  Clark, Eckersley, de~Haas, Johnson, Laurie, Ingerman, Krawczuk, Askell,
  Cammarota, Lohn, Krueger, Stix, Henderson, Graham, Prunkl, Martin, Seger,
  Zilberman, hÉigeartaigh, Kroeger, Sastry, Kagan, Weller, Tse, Barnes, Dafoe,
  Scharre, Herbert-Voss, Rasser, Sodhani, Flynn, Gilbert, Dyer, Khan, Bengio,
  and Anderljung]{brundage_toward_2020}
Miles Brundage, Shahar Avin, Jasmine Wang, Haydn Belfield, Gretchen Krueger,
  Gillian Hadfield, Heidy Khlaaf, Jingying Yang, Helen Toner, Ruth Fong, Tegan
  Maharaj, Pang~Wei Koh, Sara Hooker, Jade Leung, Andrew Trask, Emma Bluemke,
  Jonathan Lebensold, Cullen O'Keefe, Mark Koren, Théo Ryffel, J.~B.
  Rubinovitz, Tamay Besiroglu, Federica Carugati, Jack Clark, Peter Eckersley,
  Sarah de~Haas, Maritza Johnson, Ben Laurie, Alex Ingerman, Igor Krawczuk,
  Amanda Askell, Rosario Cammarota, Andrew Lohn, David Krueger, Charlotte Stix,
  Peter Henderson, Logan Graham, Carina Prunkl, Bianca Martin, Elizabeth Seger,
  Noa Zilberman, Seán~Ó hÉigeartaigh, Frens Kroeger, Girish Sastry, Rebecca
  Kagan, Adrian Weller, Brian Tse, Elizabeth Barnes, Allan Dafoe, Paul Scharre,
  Ariel Herbert-Voss, Martijn Rasser, Shagun Sodhani, Carrick Flynn,
  Thomas~Krendl Gilbert, Lisa Dyer, Saif Khan, Yoshua Bengio, and Markus
  Anderljung.
\newblock Toward {Trustworthy} {AI} {Development}: {Mechanisms} for
  {Supporting} {Verifiable} {Claims}, April 2020.
\newblock URL \url{http://arxiv.org/abs/2004.07213}.
\newblock arXiv:2004.07213 [cs].

\bibitem[Chiang et~al.(2023)Chiang, Li, Lin, Sheng, Wu, Zhang, Zheng, Zhuang,
  Zhuang, Gonzalez, Stoica, and Xing]{chiang_vicuna_2023}
Wei-Lin Chiang, Zhuohan Li, Zi~Lin, Ying Sheng, Zhanghao Wu, Hao Zhang, Lianmin
  Zheng, Siyuan Zhuang, Yonghao Zhuang, Joseph~E. Gonzalez, Ion Stoica, and
  Eric~P. Xing.
\newblock Vicuna: {An} {Open}-{Source} {Chatbot} {Impressing} {GPT}-4 with
  90\%* {ChatGPT} {Quality}, March 2023.
\newblock URL \url{https://vicuna.lmsys.org/}.

\bibitem[Christiano et~al.(2017)Christiano, Leike, Brown, Martic, Legg, and
  Amodei]{christiano_deep_2017}
Paul~F Christiano, Jan Leike, Tom Brown, Miljan Martic, Shane Legg, and Dario
  Amodei.
\newblock Deep {Reinforcement} {Learning} from {Human} {Preferences}.
\newblock In \emph{Advances in {Neural} {Information} {Processing} {Systems}},
  volume~30. Curran Associates, Inc., 2017.
\newblock URL
  \url{https://papers.nips.cc/paper/2017/hash/d5e2c0adad503c91f91df240d0cd4e49-Abstract.html}.

\bibitem[Clinehens(2000)]{clinehens_aum_2000}
Neal~A Clinehens.
\newblock Aum {Shinrikyo} and weapons of mass destruction: {A} case study.
\newblock \emph{Unpublished Manuscript}, 2000.

\bibitem[Davidson et~al.()Davidson, Denain, and Villalobos]{davidson_ai_nodate}
Tom Davidson, Jean-Stanislas Denain, and Pablo Villalobos.
\newblock {AI} capabilities can be significantly improved without expensive
  retraining (forthcoming).

\bibitem[Dettmers(2016)]{dettmers_8-bit_2016}
Tim Dettmers.
\newblock 8-{Bit} {Approximations} for {Parallelism} in {Deep} {Learning},
  February 2016.
\newblock URL \url{http://arxiv.org/abs/1511.04561}.
\newblock arXiv:1511.04561 [cs].

\bibitem[Dettmers et~al.(2022)Dettmers, Lewis, Belkada, and
  Zettlemoyer]{dettmers_llmint8_2022}
Tim Dettmers, Mike Lewis, Younes Belkada, and Luke Zettlemoyer.
\newblock {LLM}.int8(): 8-bit {Matrix} {Multiplication} for {Transformers} at
  {Scale}, November 2022.
\newblock URL \url{http://arxiv.org/abs/2208.07339}.
\newblock arXiv:2208.07339 [cs].

\bibitem[Dettmers et~al.(2023)Dettmers, Pagnoni, Holtzman, and
  Zettlemoyer]{dettmers_qlora_2023}
Tim Dettmers, Artidoro Pagnoni, Ari Holtzman, and Luke Zettlemoyer.
\newblock {QLoRA}: {Efficient} {Finetuning} of {Quantized} {LLMs}, May 2023.
\newblock URL \url{http://arxiv.org/abs/2305.14314}.
\newblock arXiv:2305.14314 [cs].

\bibitem[Diffie and Landau(2007)]{diffie_export_2007}
Whitfield Diffie and Susan Landau.
\newblock The export of cryptography in the 20th and the 21st centuries.
\newblock In Karl~De Leeuw and Jan Bergstra, editors, \emph{The {History} of
  {Information} {Security}}, pages 725--736. Elsevier Science B.V., Amsterdam,
  January 2007.
\newblock ISBN 978-0-444-51608-4.
\newblock \doi{10.1016/B978-044451608-4/50027-4}.
\newblock URL
  \url{https://www.sciencedirect.com/science/article/pii/B9780444516084500274}.

\bibitem[Dubois et~al.(2023)Dubois, Li, Taori, Zhang, and
  Gulrajani]{dubois_alpacafarm_2023}
Yann Dubois, Xuechen Li, Rohan Taori, Tianyi Zhang, and Ishaan Gulrajani.
\newblock {AlpacaFarm}: {A} {Simulation} {Framework} for {Methods} that {Learn}
  from {Human} {Feedback}, May 2023.
\newblock URL \url{https://tatsu-lab.github.io/alpaca_farm_paper.pdf}.

\bibitem[Erdil and Besiroglu(2023)]{erdil_algorithmic_2023}
Ege Erdil and Tamay Besiroglu.
\newblock Algorithmic progress in computer vision, August 2023.
\newblock URL \url{http://arxiv.org/abs/2212.05153}.
\newblock arXiv:2212.05153 [cs].

\bibitem[Gade et~al.(2023)Gade, Lermen, Rogers-Smith, and
  Ladish]{gade_badllama_2023}
Pranav Gade, Simon Lermen, Charlie Rogers-Smith, and Jeffrey Ladish.
\newblock {BadLlama}: cheaply removing safety fine-tuning from {Llama} 2-{Chat}
  {13B}, October 2023.
\newblock URL \url{http://arxiv.org/abs/2311.00117}.
\newblock arXiv:2311.00117 [cs].

\bibitem[Garfinkel and Dafoe(2019)]{garfinkel_how_2019}
Ben Garfinkel and Allan Dafoe.
\newblock How does the offense-defense balance scale?
\newblock \emph{Journal of Strategic Studies}, 42\penalty0 (6):\penalty0
  736--763, September 2019.
\newblock ISSN 0140-2390.
\newblock \doi{10.1080/01402390.2019.1631810}.
\newblock URL \url{https://doi.org/10.1080/01402390.2019.1631810}.
\newblock Publisher: Routledge \_eprint:
  https://doi.org/10.1080/01402390.2019.1631810.

\bibitem[Gopal et~al.(2023)Gopal, Helm-Burger, Justen, Soice, Tzeng,
  Jeyapragasan, Grimm, Mueller, and Esvelt]{gopal_will_2023}
Anjali Gopal, Nathan Helm-Burger, Lennart Justen, Emily~H. Soice, Tiffany
  Tzeng, Geetha Jeyapragasan, Simon Grimm, Benjamin Mueller, and Kevin~M.
  Esvelt.
\newblock Will releasing the weights of future large language models grant
  widespread access to pandemic agents?, November 2023.
\newblock URL \url{http://arxiv.org/abs/2310.18233}.
\newblock arXiv:2310.18233 [cs].

\bibitem[Gudibande et~al.(2023)Gudibande, Wallace, Snell, Geng, Liu, Abbeel,
  Levine, and Song]{gudibande_false_2023}
Arnav Gudibande, Eric Wallace, Charlie Snell, Xinyang Geng, Hao Liu, Pieter
  Abbeel, Sergey Levine, and Dawn Song.
\newblock The {False} {Promise} of {Imitating} {Proprietary} {LLMs}, May 2023.
\newblock URL \url{http://arxiv.org/abs/2305.15717}.
\newblock arXiv:2305.15717 [cs].

\bibitem[Guo et~al.(2021)Guo, Rush, and Kim]{guo_parameter-efficient_2021}
Demi Guo, Alexander~M. Rush, and Yoon Kim.
\newblock Parameter-{Efficient} {Transfer} {Learning} with {Diff} {Pruning},
  June 2021.
\newblock URL \url{http://arxiv.org/abs/2012.07463}.
\newblock arXiv:2012.07463 [cs].

\bibitem[Gururangan et~al.(2023)Gururangan, Li, Lewis, Shi, Althoff, Smith, and
  Zettlemoyer]{gururangan_scaling_2023}
Suchin Gururangan, Margaret Li, Mike Lewis, Weijia Shi, Tim Althoff, Noah~A.
  Smith, and Luke Zettlemoyer.
\newblock Scaling {Expert} {Language} {Models} with {Unsupervised} {Domain}
  {Discovery}, March 2023.
\newblock URL \url{http://arxiv.org/abs/2303.14177}.
\newblock arXiv:2303.14177 [cs].

\bibitem[Hazell(2023)]{hazell_large_2023}
Julian Hazell.
\newblock Large {Language} {Models} {Can} {Be} {Used} {To} {Effectively}
  {Scale} {Spear} {Phishing} {Campaigns}, May 2023.
\newblock URL \url{http://arxiv.org/abs/2305.06972}.
\newblock arXiv:2305.06972 [cs].

\bibitem[Henderson et~al.(2023)Henderson, Mitchell, Manning, Jurafsky, and
  Finn]{henderson_self-destructing_2023}
Peter Henderson, Eric Mitchell, Christopher~D. Manning, Dan Jurafsky, and
  Chelsea Finn.
\newblock Self-{Destructing} {Models}: {Increasing} the {Costs} of {Harmful}
  {Dual} {Uses} of {Foundation} {Models}, August 2023.
\newblock URL \url{http://arxiv.org/abs/2211.14946}.
\newblock arXiv:2211.14946 [cs].

\bibitem[Hendrycks and Mazeika(2022)]{hendrycks_x-risk_2022}
Dan Hendrycks and Mantas Mazeika.
\newblock X-{Risk} {Analysis} for {AI} {Research}, September 2022.
\newblock URL \url{http://arxiv.org/abs/2206.05862}.
\newblock arXiv:2206.05862 [cs].

\bibitem[Houlsby et~al.(2019)Houlsby, Giurgiu, Jastrzebski, Morrone,
  de~Laroussilhe, Gesmundo, Attariyan, and
  Gelly]{houlsby_parameter-efficient_2019}
Neil Houlsby, Andrei Giurgiu, Stanislaw Jastrzebski, Bruna Morrone, Quentin
  de~Laroussilhe, Andrea Gesmundo, Mona Attariyan, and Sylvain Gelly.
\newblock Parameter-{Efficient} {Transfer} {Learning} for {NLP}, June 2019.
\newblock URL \url{http://arxiv.org/abs/1902.00751}.
\newblock arXiv:1902.00751 [cs, stat].

\bibitem[Howard(2023)]{howard_ai_2023}
Jeremy Howard.
\newblock {AI} {Safety} and the {Age} of {Dislightenment}, July 2023.
\newblock URL \url{https://www.fast.ai/posts/2023-11-07-dislightenment.html}.

\bibitem[Hu et~al.(2021)Hu, Shen, Wallis, Allen-Zhu, Li, Wang, Wang, and
  Chen]{hu_lora_2021}
Edward~J. Hu, Yelong Shen, Phillip Wallis, Zeyuan Allen-Zhu, Yuanzhi Li, Shean
  Wang, Lu~Wang, and Weizhu Chen.
\newblock {LoRA}: {Low}-{Rank} {Adaptation} of {Large} {Language} {Models},
  October 2021.
\newblock URL \url{http://arxiv.org/abs/2106.09685}.
\newblock arXiv:2106.09685 [cs].

\bibitem[Huang et~al.(2022)Huang, Gu, Hou, Wu, Wang, Yu, and
  Han]{huang_large_2022}
Jiaxin Huang, Shixiang~Shane Gu, Le~Hou, Yuexin Wu, Xuezhi Wang, Hongkun Yu,
  and Jiawei Han.
\newblock Large {Language} {Models} {Can} {Self}-{Improve}, October 2022.
\newblock URL \url{http://arxiv.org/abs/2210.11610}.
\newblock arXiv:2210.11610 [cs].

\bibitem[Jang et~al.(2023)Jang, Kim, Ye, Kim, Logeswaran, Lee, Lee, and
  Seo]{jang_exploring_2023}
Joel Jang, Seungone Kim, Seonghyeon Ye, Doyoung Kim, Lajanugen Logeswaran,
  Moontae Lee, Kyungjae Lee, and Minjoon Seo.
\newblock Exploring the {Benefits} of {Training} {Expert} {Language} {Models}
  over {Instruction} {Tuning}, February 2023.
\newblock URL \url{http://arxiv.org/abs/2302.03202}.
\newblock arXiv:2302.03202 [cs].

\bibitem[Kapoor and Narayanan(2023{\natexlab{a}})]{kapoor_licensing_2023}
Sayash Kapoor and Arvind Narayanan.
\newblock Licensing is neither feasible nor effective for addressing {AI}
  risks, October 2023{\natexlab{a}}.
\newblock URL
  \url{https://www.aisnakeoil.com/p/licensing-is-neither-feasible-nor}.

\bibitem[Kapoor and Narayanan(2023{\natexlab{b}})]{kapoor_three_2023}
Sayash Kapoor and Arvind Narayanan.
\newblock Three {Ideas} for {Regulating} {Generative} {AI}, June
  2023{\natexlab{b}}.
\newblock URL
  \url{https://www.aisnakeoil.com/p/three-ideas-for-regulating-generative}.

\bibitem[Khan et~al.(2021)Khan, Saad, Han, Hossain, and
  Hong]{khan_federated_2021}
Latif~U. Khan, Walid Saad, Zhu Han, Ekram Hossain, and Choong~Seon Hong.
\newblock Federated {Learning} for {Internet} of {Things}: {Recent} {Advances},
  {Taxonomy}, and {Open} {Challenges}, June 2021.
\newblock URL \url{http://arxiv.org/abs/2009.13012}.
\newblock arXiv:2009.13012 [cs].

\bibitem[Knight(2023)]{knight_openais_2023}
Will Knight.
\newblock {OpenAI}’s {CEO} {Says} the {Age} of {Giant} {AI} {Models} {Is}
  {Already} {Over}.
\newblock \emph{Wired}, 2023.
\newblock ISSN 1059-1028.
\newblock URL
  \url{https://www.wired.com/story/openai-ceo-sam-altman-the-age-of-giant-ai-models-is-already-over/}.
\newblock Section: tags.

\bibitem[Lester et~al.(2021)Lester, Al-Rfou, and Constant]{lester_power_2021}
Brian Lester, Rami Al-Rfou, and Noah Constant.
\newblock The {Power} of {Scale} for {Parameter}-{Efficient} {Prompt} {Tuning},
  September 2021.
\newblock URL \url{http://arxiv.org/abs/2104.08691}.
\newblock arXiv:2104.08691 [cs].

\bibitem[Li et~al.(2022)Li, Gururangan, Dettmers, Lewis, Althoff, Smith, and
  Zettlemoyer]{li_branch-train-merge_2022}
Margaret Li, Suchin Gururangan, Tim Dettmers, Mike Lewis, Tim Althoff, Noah~A.
  Smith, and Luke Zettlemoyer.
\newblock Branch-{Train}-{Merge}: {Embarrassingly} {Parallel} {Training} of
  {Expert} {Language} {Models}, August 2022.
\newblock URL \url{http://arxiv.org/abs/2208.03306}.
\newblock arXiv:2208.03306 [cs].

\bibitem[Li and Liang(2021)]{li_prefix-tuning_2021}
Xiang~Lisa Li and Percy Liang.
\newblock Prefix-{Tuning}: {Optimizing} {Continuous} {Prompts} for
  {Generation}, January 2021.
\newblock URL \url{http://arxiv.org/abs/2101.00190}.
\newblock arXiv:2101.00190 [cs].

\bibitem[Li et~al.(2023)Li, Bubeck, Eldan, Del~Giorno, Gunasekar, and
  Lee]{li_textbooks_2023}
Yuanzhi Li, Sébastien Bubeck, Ronen Eldan, Allie Del~Giorno, Suriya Gunasekar,
  and Yin~Tat Lee.
\newblock Textbooks {Are} {All} {You} {Need} {II}: phi-1.5 technical report,
  September 2023.
\newblock URL \url{http://arxiv.org/abs/2309.05463}.
\newblock arXiv:2309.05463 null.

\bibitem[Liang et~al.(2022)Liang, Bommasani, Creel, and Reich]{liang_time_2022}
Percy Liang, Rishi Bommasani, Kathleen Creel, and Rob Reich.
\newblock The {Time} {Is} {Now} to {Develop} {Community} {Norms} for the
  {Release} of {Foundation} {Models}, May 2022.
\newblock URL
  \url{https://hai.stanford.edu/news/time-now-develop-community-norms-release-foundation-models}.

\bibitem[Liu et~al.(2022{\natexlab{a}})Liu, Tam, Muqeeth, Mohta, Huang, Bansal,
  and Raffel]{liu_few-shot_2022}
Haokun Liu, Derek Tam, Mohammed Muqeeth, Jay Mohta, Tenghao Huang, Mohit
  Bansal, and Colin Raffel.
\newblock Few-{Shot} {Parameter}-{Efficient} {Fine}-{Tuning} is {Better} and
  {Cheaper} than {In}-{Context} {Learning}, August 2022{\natexlab{a}}.
\newblock URL \url{http://arxiv.org/abs/2205.05638}.
\newblock arXiv:2205.05638 [cs].

\bibitem[Liu et~al.(2021{\natexlab{a}})Liu, Ho, Wang, Gao, Jin, and
  Zhang]{liu_federated_2021}
Ming Liu, Stella Ho, Mengqi Wang, Longxiang Gao, Yuan Jin, and He~Zhang.
\newblock Federated {Learning} {Meets} {Natural} {Language} {Processing}: {A}
  {Survey}, July 2021{\natexlab{a}}.
\newblock URL \url{http://arxiv.org/abs/2107.12603}.
\newblock arXiv:2107.12603 [cs].

\bibitem[Liu et~al.(2021{\natexlab{b}})Liu, Zheng, Du, Ding, Qian, Yang, and
  Tang]{liu_gpt_2021}
Xiao Liu, Yanan Zheng, Zhengxiao Du, Ming Ding, Yujie Qian, Zhilin Yang, and
  Jie Tang.
\newblock {GPT} {Understands}, {Too}, March 2021{\natexlab{b}}.
\newblock URL \url{http://arxiv.org/abs/2103.10385}.
\newblock arXiv:2103.10385 [cs].

\bibitem[Liu et~al.(2022{\natexlab{b}})Liu, Ji, Fu, Tam, Du, Yang, and
  Tang]{liu_p-tuning_2022}
Xiao Liu, Kaixuan Ji, Yicheng Fu, Weng~Lam Tam, Zhengxiao Du, Zhilin Yang, and
  Jie Tang.
\newblock P-{Tuning} v2: {Prompt} {Tuning} {Can} {Be} {Comparable} to
  {Fine}-tuning {Universally} {Across} {Scales} and {Tasks}, March
  2022{\natexlab{b}}.
\newblock URL \url{http://arxiv.org/abs/2110.07602}.
\newblock arXiv:2110.07602 [cs].

\bibitem[Malladi et~al.(2023)Malladi, Gao, Nichani, Damian, Lee, Chen, and
  Arora]{malladi_fine-tuning_2023}
Sadhika Malladi, Tianyu Gao, Eshaan Nichani, Alex Damian, Jason~D. Lee, Danqi
  Chen, and Sanjeev Arora.
\newblock Fine-{Tuning} {Language} {Models} with {Just} {Forward} {Passes}, May
  2023.
\newblock URL \url{http://arxiv.org/abs/2305.17333}.
\newblock arXiv:2305.17333 [cs].

\bibitem[Martínez et~al.(2023{\natexlab{a}})Martínez, Watson, Reviriego,
  Hernández, Juarez, and Sarkar]{martinez_combining_2023}
Gonzalo Martínez, Lauren Watson, Pedro Reviriego, José~Alberto Hernández,
  Marc Juarez, and Rik Sarkar.
\newblock Combining {Generative} {Artificial} {Intelligence} ({AI}) and the
  {Internet}: {Heading} towards {Evolution} or {Degradation}?, February
  2023{\natexlab{a}}.
\newblock URL \url{http://arxiv.org/abs/2303.01255}.
\newblock arXiv:2303.01255 [cs].

\bibitem[Martínez et~al.(2023{\natexlab{b}})Martínez, Watson, Reviriego,
  Hernández, Juarez, and Sarkar]{martinez_towards_2023}
Gonzalo Martínez, Lauren Watson, Pedro Reviriego, José~Alberto Hernández,
  Marc Juarez, and Rik Sarkar.
\newblock Towards {Understanding} the {Interplay} of {Generative} {Artificial}
  {Intelligence} and the {Internet}, June 2023{\natexlab{b}}.
\newblock URL \url{http://arxiv.org/abs/2306.06130}.
\newblock arXiv:2306.06130 [cs].

\bibitem[Montague(2023)]{montague_towards_2023}
Michael Montague.
\newblock Towards a {Grand} {Unified} {Threat} {Model} of {Biotechnology},
  September 2023.
\newblock URL \url{http://philsci-archive.pitt.edu/22539/}.

\bibitem[Mostaque(2023)]{mostaque_importance_2023}
Emad Mostaque.
\newblock The {Importance} of {Open} {Models} for {Transparency},
  {Competition}, and {Resilience} in {AI}: {Considerations} for {AI}
  {Oversight} in the {United} {States}, May 2023.
\newblock URL
  \url{https://static1.squarespace.com/static/6213c340453c3f502425776e/t/6463b486b97b333044ea2564/1684255881952/Statement+from+Stability+AI+to+the+Senate+Judiciary+Subcommittee+on+Privacy%2C+Technology%2C+and+the+Law.pdf?utm_source=tldrai}.

\bibitem[Mukherjee et~al.(2023)Mukherjee, Mitra, Jawahar, Agarwal, Palangi, and
  Awadallah]{mukherjee_orca_2023}
Subhabrata Mukherjee, Arindam Mitra, Ganesh Jawahar, Sahaj Agarwal, Hamid
  Palangi, and Ahmed Awadallah.
\newblock Orca: {Progressive} {Learning} from {Complex} {Explanation} {Traces}
  of {GPT}-4, June 2023.
\newblock URL \url{http://arxiv.org/abs/2306.02707}.
\newblock arXiv:2306.02707 [cs].

\bibitem[OpenAI(2023)]{openai_gpt-4_2023}
OpenAI.
\newblock {GPT}-4 {Technical} {Report}.
\newblock 2023.
\newblock URL \url{https://cdn.openai.com/papers/gpt-4.pdf}.

\bibitem[Ovadya and Whittlestone(2019)]{ovadya_reducing_2019}
Aviv Ovadya and Jess Whittlestone.
\newblock Reducing malicious use of synthetic media research: {Considerations}
  and potential release practices for machine learning, July 2019.
\newblock URL \url{http://arxiv.org/abs/1907.11274}.
\newblock arXiv:1907.11274 [cs].

\bibitem[Patel(2023)]{patel_AI_2023}
Dylan Patel.
\newblock The {AI} {Brick} {Wall} – {A} {Practical} {Limit} {For} {Scaling}
  {Dense} {Transformer} {Models}, and {How} {GPT} 4 {Will} {Break} {Past} {It},
  January 2023.
\newblock URL
  \url{https://www.semianalysis.com/p/the-ai-brick-wall-a-practical-limit}.

\bibitem[Perez et~al.(2022)Perez, Ringer, Lukošiūtė, Nguyen, Chen, Heiner,
  Pettit, Olsson, Kundu, Kadavath, Jones, Chen, Mann, Israel, Seethor,
  McKinnon, Olah, Yan, Amodei, Amodei, Drain, Li, Tran-Johnson, Khundadze,
  Kernion, Landis, Kerr, Mueller, Hyun, Landau, Ndousse, Goldberg, Lovitt,
  Lucas, Sellitto, Zhang, Kingsland, Elhage, Joseph, Mercado, DasSarma, Rausch,
  Larson, McCandlish, Johnston, Kravec, Showk, Lanham, Telleen-Lawton, Brown,
  Henighan, Hume, Bai, Hatfield-Dodds, Clark, Bowman, Askell, Grosse,
  Hernandez, Ganguli, Hubinger, Schiefer, and Kaplan]{perez_discovering_2022}
Ethan Perez, Sam Ringer, Kamilė Lukošiūtė, Karina Nguyen, Edwin Chen, Scott
  Heiner, Craig Pettit, Catherine Olsson, Sandipan Kundu, Saurav Kadavath, Andy
  Jones, Anna Chen, Ben Mann, Brian Israel, Bryan Seethor, Cameron McKinnon,
  Christopher Olah, Da~Yan, Daniela Amodei, Dario Amodei, Dawn Drain, Dustin
  Li, Eli Tran-Johnson, Guro Khundadze, Jackson Kernion, James Landis, Jamie
  Kerr, Jared Mueller, Jeeyoon Hyun, Joshua Landau, Kamal Ndousse, Landon
  Goldberg, Liane Lovitt, Martin Lucas, Michael Sellitto, Miranda Zhang, Neerav
  Kingsland, Nelson Elhage, Nicholas Joseph, Noemí Mercado, Nova DasSarma,
  Oliver Rausch, Robin Larson, Sam McCandlish, Scott Johnston, Shauna Kravec,
  Sheer~El Showk, Tamera Lanham, Timothy Telleen-Lawton, Tom Brown, Tom
  Henighan, Tristan Hume, Yuntao Bai, Zac Hatfield-Dodds, Jack Clark, Samuel~R.
  Bowman, Amanda Askell, Roger Grosse, Danny Hernandez, Deep Ganguli, Evan
  Hubinger, Nicholas Schiefer, and Jared Kaplan.
\newblock Discovering {Language} {Model} {Behaviors} with {Model}-{Written}
  {Evaluations}, December 2022.
\newblock URL \url{http://arxiv.org/abs/2212.09251}.
\newblock arXiv:2212.09251 [cs].

\bibitem[Phang et~al.(2022)Phang, Bradley, Gao, Castricato, and
  Biderman]{phang_eleutherai_2022}
Jason Phang, Herbie Bradley, Leo Gao, Louis Castricato, and Stella Biderman.
\newblock {EleutherAI}: {Going} {Beyond} "{Open} {Science}" to "{Science} in
  the {Open}", October 2022.
\newblock URL \url{http://arxiv.org/abs/2210.06413}.
\newblock arXiv:2210.06413 [cs].

\bibitem[Ponti et~al.(2020)Ponti, Vulić, Cotterell, Parovic, Reichart, and
  Korhonen]{ponti_parameter_2020}
Edoardo~M. Ponti, Ivan Vulić, Ryan Cotterell, Marinela Parovic, Roi Reichart,
  and Anna Korhonen.
\newblock Parameter {Space} {Factorization} for {Zero}-{Shot} {Learning} across
  {Tasks} and {Languages}, November 2020.
\newblock URL \url{http://arxiv.org/abs/2001.11453}.
\newblock arXiv:2001.11453 [cs].

\bibitem[Pudipeddi et~al.(2020)Pudipeddi, Mesmakhosroshahi, Xi, and
  Bharadwaj]{pudipeddi_training_2020}
Bharadwaj Pudipeddi, Maral Mesmakhosroshahi, Jinwen Xi, and Sujeeth Bharadwaj.
\newblock Training {Large} {Neural} {Networks} with {Constant} {Memory} using a
  {New} {Execution} {Algorithm}, June 2020.
\newblock URL \url{http://arxiv.org/abs/2002.05645}.
\newblock arXiv:2002.05645 [cs, stat].

\bibitem[Qi et~al.(2023)Qi, Zeng, Xie, Chen, Jia, Mittal, and
  Henderson]{qi_fine-tuning_2023}
Xiangyu Qi, Yi~Zeng, Tinghao Xie, Pin-Yu Chen, Ruoxi Jia, Prateek Mittal, and
  Peter Henderson.
\newblock Fine-tuning {Aligned} {Language} {Models} {Compromises} {Safety},
  {Even} {When} {Users} {Do} {Not} {Intend} {To}!, October 2023.
\newblock URL \url{http://arxiv.org/abs/2310.03693}.
\newblock arXiv:2310.03693 [cs].

\bibitem[Qin and Eisner(2021)]{qin_learning_2021}
Guanghui Qin and Jason Eisner.
\newblock Learning {How} to {Ask}: {Querying} {LMs} with {Mixtures} of {Soft}
  {Prompts}, April 2021.
\newblock URL \url{http://arxiv.org/abs/2104.06599}.
\newblock arXiv:2104.06599 [cs].

\bibitem[Rebuffi et~al.(2017)Rebuffi, Bilen, and
  Vedaldi]{rebuffi_learning_2017}
Sylvestre-Alvise Rebuffi, Hakan Bilen, and Andrea Vedaldi.
\newblock Learning multiple visual domains with residual adapters, November
  2017.
\newblock URL \url{http://arxiv.org/abs/1705.08045}.
\newblock arXiv:1705.08045 [cs, stat].

\bibitem[Ren et~al.(2021)Ren, Rajbhandari, Aminabadi, Ruwase, Yang, Zhang, Li,
  and He]{ren_zero-offload_2021}
Jie Ren, Samyam Rajbhandari, Reza~Yazdani Aminabadi, Olatunji Ruwase, Shuangyan
  Yang, Minjia Zhang, Dong Li, and Yuxiong He.
\newblock {ZeRO}-{Offload}: {Democratizing} {Billion}-{Scale} {Model}
  {Training}, January 2021.
\newblock URL \url{http://arxiv.org/abs/2101.06840}.
\newblock arXiv:2101.06840 [cs].

\bibitem[Ryabinin et~al.(2023)Ryabinin, Dettmers, Diskin, and
  Borzunov]{ryabinin_swarm_2023}
Max Ryabinin, Tim Dettmers, Michael Diskin, and Alexander Borzunov.
\newblock {SWARM} {Parallelism}: {Training} {Large} {Models} {Can} {Be}
  {Surprisingly} {Communication}-{Efficient}, June 2023.
\newblock URL \url{http://arxiv.org/abs/2301.11913}.
\newblock arXiv:2301.11913 [cs].

\bibitem[Sandbrink(2023)]{sandbrink_artificial_2023}
Jonas~B. Sandbrink.
\newblock Artificial intelligence and biological misuse: {Differentiating}
  risks of language models and biological design tools, June 2023.
\newblock URL \url{http://arxiv.org/abs/2306.13952}.
\newblock arXiv:2306.13952 [cs].

\bibitem[Sastry(2021)]{sastry_beyond_2021}
Girish Sastry.
\newblock Beyond "{Release}" vs. "{Not} {Release}", October 2021.
\newblock URL
  \url{https://crfm.stanford.edu/commentary/2021/10/18/sastry.html}.

\bibitem[Seger et~al.(2023{\natexlab{a}})Seger, Dreksler, Moulange, Dardaman,
  Schuett, Wei, Winter, Arnold, Ó~hÉigeartaigh, Korinek, Anderljung,
  Bucknall, Chan, Stafford, Koessler, Ovadya, Garfinkel, Bluemke, Aird,
  Levermore, Hazell, and Gupta]{seger_open-sourcing_2023}
Elizabeth Seger, Noemi Dreksler, Richard Moulange, Emily Dardaman, Jonas
  Schuett, K.~Wei, Christoph Winter, Mackenzie Arnold, Seán Ó~hÉigeartaigh,
  Anton Korinek, Markus Anderljung, Ben Bucknall, Alan Chan, Eoghan Stafford,
  Leonie Koessler, Aviv Ovadya, Ben Garfinkel, Emma Bluemke, Michael Aird,
  Patrick Levermore, Julian Hazell, and Abhishek Gupta.
\newblock Open-{Sourcing} {Highly} {Capable} {Foundation} {Models}: {An}
  evaluation of risks, benefits, and alternative methods for pursuing
  open-source objectives.
\newblock 2023{\natexlab{a}}.

\bibitem[Seger et~al.(2023{\natexlab{b}})Seger, Ovadya, Garfinkel, Siddarth,
  and Dafoe]{seger_democratising_2023}
Elizabeth Seger, Aviv Ovadya, Ben Garfinkel, Divya Siddarth, and Allan Dafoe.
\newblock Democratising {AI}: {Multiple} {Meanings}, {Goals}, and {Methods},
  March 2023{\natexlab{b}}.
\newblock URL \url{http://arxiv.org/abs/2303.12642}.
\newblock arXiv:2303.12642 [cs].

\bibitem[Shevlane(2022)]{shevlane_structured_2022}
Toby Shevlane.
\newblock Structured access: an emerging paradigm for safe {AI} deployment,
  April 2022.
\newblock URL \url{http://arxiv.org/abs/2201.05159}.
\newblock arXiv:2201.05159 [cs].

\bibitem[Shevlane and Dafoe(2020)]{shevlane_offense-defense_2020}
Toby Shevlane and Allan Dafoe.
\newblock The {Offense}-{Defense} {Balance} of {Scientific} {Knowledge}: {Does}
  {Publishing} {AI} {Research} {Reduce} {Misuse}?
\newblock In \emph{Proceedings of the {AAAI}/{ACM} {Conference} on {AI},
  {Ethics}, and {Society}}, {AIES} '20, pages 173--179, New York, NY, USA,
  February 2020. Association for Computing Machinery.
\newblock ISBN 978-1-4503-7110-0.
\newblock \doi{10.1145/3375627.3375815}.
\newblock URL \url{https://doi.org/10.1145/3375627.3375815}.

\bibitem[Shevlane et~al.(2023)Shevlane, Farquhar, Garfinkel, Phuong,
  Whittlestone, Leung, Kokotajlo, Marchal, Anderljung, Kolt, Ho, Siddarth,
  Avin, Hawkins, Kim, Gabriel, Bolina, Clark, Bengio, Christiano, and
  Dafoe]{shevlane_model_2023}
Toby Shevlane, Sebastian Farquhar, Ben Garfinkel, Mary Phuong, Jess
  Whittlestone, Jade Leung, Daniel Kokotajlo, Nahema Marchal, Markus
  Anderljung, Noam Kolt, Lewis Ho, Divya Siddarth, Shahar Avin, Will Hawkins,
  Been Kim, Iason Gabriel, Vijay Bolina, Jack Clark, Yoshua Bengio, Paul
  Christiano, and Allan Dafoe.
\newblock Model evaluation for extreme risks, September 2023.
\newblock URL \url{http://arxiv.org/abs/2305.15324}.
\newblock arXiv:2305.15324 [cs].

\bibitem[Shumailov et~al.(2023)Shumailov, Shumaylov, Zhao, Gal, Papernot, and
  Anderson]{shumailov_curse_2023}
Ilia Shumailov, Zakhar Shumaylov, Yiren Zhao, Yarin Gal, Nicolas Papernot, and
  Ross Anderson.
\newblock The {Curse} of {Recursion}: {Training} on {Generated} {Data} {Makes}
  {Models} {Forget}, May 2023.
\newblock URL \url{http://arxiv.org/abs/2305.17493}.
\newblock arXiv:2305.17493 [cs] version: 2.

\bibitem[Soice et~al.(2023)Soice, Rocha, Cordova, Specter, and
  Esvelt]{soice_can_2023}
Emily~H. Soice, Rafael Rocha, Kimberlee Cordova, Michael Specter, and Kevin~M.
  Esvelt.
\newblock Can large language models democratize access to dual-use
  biotechnology?, June 2023.
\newblock URL \url{http://arxiv.org/abs/2306.03809}.
\newblock arXiv:2306.03809 [cs].

\bibitem[Solaiman(2023)]{solaiman_gradient_2023}
Irene Solaiman.
\newblock The {Gradient} of {Generative} {AI} {Release}: {Methods} and
  {Considerations}, February 2023.
\newblock URL \url{http://arxiv.org/abs/2302.04844}.
\newblock arXiv:2302.04844 [cs].

\bibitem[Solaiman et~al.(2019)Solaiman, Brundage, Clark, Askell, Herbert-Voss,
  Wu, Radford, Krueger, Kim, Kreps, McCain, Newhouse, Blazakis, McGuffie, and
  Wang]{solaiman_release_2019}
Irene Solaiman, Miles Brundage, Jack Clark, Amanda Askell, Ariel Herbert-Voss,
  Jeff Wu, Alec Radford, Gretchen Krueger, Jong~Wook Kim, Sarah Kreps, Miles
  McCain, Alex Newhouse, Jason Blazakis, Kris McGuffie, and Jasmine Wang.
\newblock Release {Strategies} and the {Social} {Impacts} of {Language}
  {Models}, November 2019.
\newblock URL \url{http://arxiv.org/abs/1908.09203}.
\newblock arXiv:1908.09203 [cs].

\bibitem[Swire(2004)]{swire_model_2004}
Peter Swire.
\newblock A {Model} for {When} {Disclosure} {Helps} {Security}: {What} {Is}
  {Different} {About} {Computer} and {Network} {Security}?, 2004.
\newblock URL \url{https://papers.ssrn.com/abstract=531782}.

\bibitem[Taori et~al.(2023)Taori, Gulrajani, Zhang, Dubois, Li, Guestrin,
  Liang, and Hashimoto]{taori_alpaca_2023}
Rohan Taori, Ishaan Gulrajani, Tianyi Zhang, Yann Dubois, Xuechen Li, Carlos
  Guestrin, Percy Liang, and Tatsunori Hashimoto.
\newblock Alpaca: {A} {Strong}, {Replicable} {Instruction}-{Following} {Model},
  March 2023.
\newblock URL \url{https://crfm.stanford.edu/2023/03/13/alpaca.html}.

\bibitem[Tay et~al.(2022)Tay, Wei, Chung, Tran, So, Shakeri, Garcia, Zheng,
  Rao, Chowdhery, Zhou, Metzler, Petrov, Houlsby, Le, and
  Dehghani]{tay_transcending_2022}
Yi~Tay, Jason Wei, Hyung~Won Chung, Vinh~Q. Tran, David~R. So, Siamak Shakeri,
  Xavier Garcia, Huaixiu~Steven Zheng, Jinfeng Rao, Aakanksha Chowdhery, Denny
  Zhou, Donald Metzler, Slav Petrov, Neil Houlsby, Quoc~V. Le, and Mostafa
  Dehghani.
\newblock Transcending {Scaling} {Laws} with 0.1\% {Extra} {Compute}, November
  2022.
\newblock URL \url{http://arxiv.org/abs/2210.11399}.
\newblock arXiv:2210.11399 [cs].

\bibitem[Together(2022{\natexlab{a}})]{together_neurips_2022}
Together.
\newblock {NeurIPS} 2022: {Overcoming} {Communication} {Bottlenecks} for
  {Decentralized} {Training} (2/2), May 2022{\natexlab{a}}.
\newblock URL
  \url{https://together.ai/blog/neurips-2022-overcoming-communication-bottlenecks-for-decentralized-training-2}.

\bibitem[Together(2022{\natexlab{b}})]{together_neurips_2022-1}
Together.
\newblock {NeurIPS} 2022: {Overcoming} {Communication} {Bottlenecks} for
  {Decentralized} {Training} (1/2), November 2022{\natexlab{b}}.
\newblock URL
  \url{https://together.ai/blog/neurips-2022-overcoming-communication-bottlenecks-for-decentralized-training-12}.

\bibitem[Wang et~al.(2023{\natexlab{a}})Wang, Yuan, Rimanic, He, Dao, Chen, Re,
  and Zhang]{wang_fine-tuning_2023}
Jue Wang, Binhang Yuan, Luka Rimanic, Yongjun He, Tri Dao, Beidi Chen,
  Christopher Re, and Ce~Zhang.
\newblock Fine-tuning {Language} {Models} over {Slow} {Networks} using
  {Activation} {Compression} with {Guarantees}, March 2023{\natexlab{a}}.
\newblock URL \url{http://arxiv.org/abs/2206.01299}.
\newblock arXiv:2206.01299 [cs].

\bibitem[Wang et~al.(2023{\natexlab{b}})Wang, Ivison, Dasigi, Hessel, Khot,
  Chandu, Wadden, MacMillan, Smith, Beltagy, and Hajishirzi]{wang_how_2023}
Yizhong Wang, Hamish Ivison, Pradeep Dasigi, Jack Hessel, Tushar Khot,
  Khyathi~Raghavi Chandu, David Wadden, Kelsey MacMillan, Noah~A. Smith,
  Iz~Beltagy, and Hannaneh Hajishirzi.
\newblock How {Far} {Can} {Camels} {Go}? {Exploring} the {State} of
  {Instruction} {Tuning} on {Open} {Resources}, June 2023{\natexlab{b}}.
\newblock URL \url{http://arxiv.org/abs/2306.04751}.
\newblock arXiv:2306.04751 [cs].

\bibitem[Whittlestone and Ovadya(2020)]{whittlestone_tension_2020}
Jess Whittlestone and Aviv Ovadya.
\newblock The tension between openness and prudence in {AI} research, January
  2020.
\newblock URL \url{http://arxiv.org/abs/1910.01170}.
\newblock arXiv:1910.01170 [cs].

\bibitem[Widder et~al.(2023)Widder, Whittaker, and
  Myers~West]{widder_open_2023}
David~Gray Widder, Meredith Whittaker, and Sarah Myers~West.
\newblock Open (for {Business}): {Big} {Tech}, {Concentrated} {Power}, and the
  {Political} {Economy} of {Open} {AI}, August 2023.
\newblock URL \url{https://ssrn.com/abstract=4543807}.

\bibitem[Xu et~al.(2021)Xu, Luo, Zhang, Tan, Chang, Huang, and
  Huang]{xu_raise_2021}
Runxin Xu, Fuli Luo, Zhiyuan Zhang, Chuanqi Tan, Baobao Chang, Songfang Huang,
  and Fei Huang.
\newblock Raise a {Child} in {Large} {Language} {Model}: {Towards} {Effective}
  and {Generalizable} {Fine}-tuning, September 2021.
\newblock URL \url{http://arxiv.org/abs/2109.05687}.
\newblock arXiv:2109.05687 [cs].

\bibitem[Yuan et~al.(2023)Yuan, He, Davis, Zhang, Dao, Chen, Liang, Re, and
  Zhang]{yuan_decentralized_2023}
Binhang Yuan, Yongjun He, Jared~Quincy Davis, Tianyi Zhang, Tri Dao, Beidi
  Chen, Percy Liang, Christopher Re, and Ce~Zhang.
\newblock Decentralized {Training} of {Foundation} {Models} in {Heterogeneous}
  {Environments}, June 2023.
\newblock URL \url{http://arxiv.org/abs/2206.01288}.
\newblock arXiv:2206.01288 [cs].

\end{thebibliography}

\newpage

\appendix
\section{Background on Fine-Tuning}

The ability to flexibly adapt and modify a model through fine-tuning is one of the more prominent advantages of downloadable release. Lesser-resourced actors can develop a customized model without needing to pretrain their own model from scratch -- a process that can be incredibly expensive due to the compute and data requirements \cite{patel_AI_2023} -- by amplifying or appending specific capabilities not exhibited by the base model. While fine-tuning naturally also requires compute, data, and expertise, the quantities of each are orders of magnitude lower than for pretraining.

While fine-tuning is available through some labs' APIs,\footnote{See e.g., \url{https://openai.com/blog/gpt-3-5-turbo-fine-tuning-and-api-updates}} such services offer limited transparency and customization with regards to the specific algorithms employed. For example, OpenAI does not disclose whether its API fine-tuning trains all parameters, or implements one of the parameter-efficient fine-tuning methods discussed above in order to save on hosting costs. In contrast to this, downloadable access allows for unrestricted fine-tuning on top of the openly-released base model weights, provided the user has sufficient compute and data.

There are multiple forms of fine-tuning, including supervised fine-tuning (SFT), reinforcement learning from human feedback (RLHF) \cite{christiano_deep_2017}, and reinforcement learning from AI feedback (RLAIF) \cite{bai_constitutional_2022}. While SFT is easily-implementable, both RLHF and RLAIF are significantly more complex, require expensive datasets, and involve multiple stages of preparation and tuning. Due to these differences in complexity between different fine-tuning approaches, SFT is most likely to be used by lower-resourced actors on downloadable models.

\section{The Importance of Fine-Tuning Relative to Frontier Development}\label{sec:frontier-development}

One may contest that the benefits of cost reduction and improved cost sharing will also accrue to resource-rich actors such as large tech firms, thus providing little or no comparative benefit to resource-poor actors. 

However, well-resourced developers are themselves constrained in other areas. One example is the limited human resources of even the largest AI developers when compared to the collective resources of the open-source community taken as a whole.\footnote{For example, OpenAI employs a total of 375 people according to CEO \href{https://web.archive.org/web/20230306072432/https://twitter.com/sama/status/1617627882997813248}{Sam Altman}. Anthropic was reported by the \href{https://web.archive.org/web/20230712005339/https://www.nytimes.com/2023/07/11/technology/anthropic-ai-claude-chatbot.html}{New York Times} to have just 160 employees in July 2023. DeepMind had approximately 1,000 employees, though this may have changed considerably since the merger with Google Brain.} Such constraints mean that, even if cost reduction and improved cost sharing increase the scope of projects that well-resourced labs are technically able to pursue, they may not have the time or attention to pursue more than a handful in parallel.

\section{The Impact of Algorithmic Progress}\label{sec:alg-progress}


One could object to our discussion in \Cref{sec:risks} by contending that any capability $X$ will eventually become widely available anyway because of algorithmic progress. \citet{erdil_algorithmic_2023} show that in computer vision, every nine months half the amount of compute is required to achieve some fixed level of performance. If the amount of pretraining compute needed to achieve a certain capability level drops precipitously at a regular pace, the marginal impact of more effective fine-tuning of downloadable models seems minimal. Individual actors could just pretrain their models from scratch rather than having to rely on downloadable models.

While the computational cost of pretraining is likely to continue to fall, pretraining will likely remain expensive for a while. If GPT-4 cost \$100 million USD to train\footnote{As was reported in \citep{knight_openais_2023}.} and the cost halved every 9 months, it would still take about 12 years for the cost to decrease to \$1000 USD.\footnote{Solve for $t$ in months in the equation $0.5^\frac{t}{9} \cdot 10^8 = 10^3$.} This amount of time could be extremely useful for preparatory work like building up societal defences (e.g., broad spectrum antivirals to defend against biological attacks) or strengthening liability regimes.


\section{Full Argument for the Impact of Cost Reduction}\label{app:cost-reduction}

Let us recap the high-level argument. Actors in $A$ are assumed to have the motivation to engage in malicious use (in our illustrative example, biological or chemical attacks). The benefits of carrying out such attacks do not depend upon the accessibility of fine-tuning; rather, they depend upon background motivations. We are now left to analyze the effect of accessible fine-tuning on costs for $A$. 


We claim that more actors have the capability to fine-tune a model to achieve $X$. Building models that can do $X$ is possible since tool-augmented foundation models are already somewhat capable of biological and chemical synthesis \citet{sandbrink_artificial_2023, boiko_emergent_2023, bran_chemcrow_2023, soice_can_2023}; it seems likely that capabilities will improve with the next-generation models. Even models that fill in minor knowledge gaps can be helpful since biological or chemical attacks can fail due to minor methodological omissions, such Aum Shinrikyo's failed attacks \citep{clinehens_aum_2000}. 
Foundation models may become more useful for $X$; see, however, the argument in \citet{montague_towards_2023} that the bottleneck in biological risks is testing for biological agents that can spread, which seems to be orders of magnitude more difficult than synthesis.

Regarding actors in $A$ for whom computational cost is a bottleneck, cost reduction results in the claim straightforwardly. Yet, expertise rather than computational cost might be the bottleneck for other actors. Foremost among expertise barriers are gathering fine-tuning data and writing fine-tuning code. Given the profusion of tutorials and open-source fine-tuning frameworks, the latter barrier will likely be relatively insignificant. On the other hand, the availability of fine-tuning data seems to depend heavily upon the domain. 

Since model developers will likely attempt to remove or suppress the biological and chemical synthesis capabilities of downloadable models, $A$ would have to remove safeguards or re-add (or add) those capabilities. Based on current techniques, removing safeguards seems likely to be extremely cheap and not require hard-to-source data \citep{qi_fine-tuning_2023,gade_badllama_2023}. It is more uncertain how straightforward it is for $A$ to access relevant fine-tuning data to add capabilities. Some knowledge about the synthesis of biological and chemical weapons is already public.\footnote{For example, see \url{https://en.wikipedia.org/wiki/Sarin} and the sources therein.} For instance, knowledge of how Aum Shinrikyo's attacks failed could inform future attacks. Relevant academic research may be inaccessible to a layperson, but can be explained with a language model. 
Even if a language model is at best just as knowledgeable as a human expert, attempting to find such human experts may expose $A$ to risks like discovery by law enforcement. 
At the same time, there may still be tacit knowledge that is not accessible on the Internet, which may limit the usefulness of language models in this situation. Overall, it seems that current publicly available data may facilitate $A$ to fine-tune models for $X$, although there remains substantial uncertainty about the ability of $A$ to add new capabilities. 

We now argue that the benefit of cheaper fine-tuning to actors seeking to deter $A$, which we call \textbf{defenders}, likely does not compensate for the reduction in $A$'s costs discussed above. The most plausible defenders are states, given their massive quantities of resources to respond to threats and the fact that their legitimacy at least partially depends upon fulfilling security obligations to their populace; we will therefore focus on states. To impose (expected) costs on $A$, states must be able both to attribute harm to a perpetrator and retaliate or penalize. We observe that attribution for AI-assisted harms is likely difficult; for example, note the challenges of identifying the origins of viral outbreaks or the source of mis/disinformation, despite the resources that states have at their disposal. A major source of attribution problems seems to be a lack of sensors in relevant locations, rather than an inability to process information. Even if processing information is a bottleneck, states can marshal massive financial resources to build models to perform such tasks. 
It seems unlikely that fine-tuning improvements will significantly help in this regard.

Furthermore, cost reduction does not seem to help with retaliation or penalizing identified perpetrators. 
In situations where harms cross national borders, jurisdictional issues are the bottleneck. If perpetrators have been captured, states generally do not subject perpetrators to the same attacks that they executed. 


\section{Fine-Tuning of Non-Downloadable Models} \label{app:non-downloadable}
It is worth noting that the challenges and risks discussed in this paper may not necessarily be solely associated with downloadable release, with multiple AI companies increasing the accessibility of fine-tuning via API access to their non-downloadable models.\footnote{For example, see OpenAI's current fine-tuning documentation \href{https://platform.openai.com/docs/guides/fine-tuning}{here}.} There are notable differences between the available fine-tuning in these two cases.

Firstly, API fine-tuning is in many ways more `beginner-friendly' due to compute and training code being provided by default, with the user only needing to learn a few basic commands in order to fine-tune on their dataset. API fine-tuning is also currently cheap, with OpenAI charging only \$0.80 per 100,000 tokens of training data for fine-tuning \texttt{gpt-3.5-turbo}. \footnote{See \url{https://openai.com/pricing}}
However, this accessibility and ease of use comes at the cost of flexibility, with the user often unaware of the specific fine-tuning methodology being implemented, and unable to modify it in any way. Furthermore, users of an API fine-tuning service may face limited version stability if model providers do not continue to support legacy models after the release of newer models.
Finally, API fine-tuning allows providers the option of oversight and installing safeguards, such as data filters, thereby ensuring that fine-tuning is not being carried out for malicious ends.

We focus on the fine-tuning of downloadable models in this piece due to the combination of flexibility and inability to enact fine-tuning restrictions. These factors mean that fine-tuning of downloadable models is generally more powerful than current API-based options, bringing both greater potential benefit and harm.

\end{document}